\documentclass[10pt,letterpaper]{article}

\usepackage{cogsci}
\usepackage{graphicx}
\usepackage{amsmath}
\usepackage{amsfonts}
\usepackage{algorithm}
\usepackage{algpseudocode}
\usepackage{url}

\cogscifinalcopy % Uncomment this line for the final submission 

\usepackage{pslatex}
\usepackage{float} % Roger Levy added this and changed figure/table
\title{Cognitive maps are generative programs}

\author{ Marta Kryven (marta.kryven@dal.ca) \\
  Faculty of Computer Science \\
  Dalhousie University
  \And  Cole Wyeth (cwyeth@uwaterloo.ca) \\
  Department of Computer Science \\
  University of Waterloo
  \AND {Aidan Curtis (curtisa@mit.edu)} \\
  Computer Science and Artificial Intelligence Laboratory \\
  Massachussetts Institute of Technology \\ 
  \And Kevin Ellis (kellis@cornell.edu) \\
  Department of Computer Science \\
  Cornell University}

\begin{document}

\maketitle

\begin{abstract}
Making sense of the world and acting in it relies on building simplified mental representations that abstract away aspects of reality. This principle of cognitive mapping is universal to agents with limited resources. Living organisms, people, and algorithms all face the problem of forming functional representations of their world under various computing constraints. In this work, we explore the hypothesis that human resource-efficient planning may arise from representing the world as predictably structured. Building on the metaphor of concepts as programs, we propose that cognitive maps can take the form of generative programs that exploit predictability and redundancy, in contrast to directly encoding spatial layouts. We use a behavioral experiment to show that people who navigate in structured spaces rely on modular planning strategies that align with programmatic map representations. %Further, the extent of modular planning is correlated with individual cognitive resources, suggesting that programmatic cognitive maps are resource-efficient adaptations.
We describe a computational model that predicts human behavior in a variety of structured scenarios. This model infers a small distribution over possible programmatic cognitive maps conditioned on human prior knowledge of the world, and uses this distribution to generate resource-efficient plans. Our models leverages a Large Language Model as an embedding of human priors, implicitly learned through training on a vast corpus of human data. Our model demonstrates improved computational efficiency, requires drastically less memory, and outperforms unstructured planning algorithms with cognitive constraints at predicting human behavior, suggesting that human planning strategies rely on programmatic cognitive maps.

\end{abstract}

\section{Introduction}

% same content as below, but trying to rewrite this as a reaserch question, and secondarily an engineering question

%People solve real-world planning problems intractable to modern AI algorithms with relative ease. For example, we efficiently navigate cities without knowing exact location of every link in the street network (Fig.~\ref{fig1}a) \cite{bongiorno2021vectorbased}, and plan construction projects with thousands of actions (Fig.~\ref{fig1}b). What underlies this efficiency, and what steps are needed to narrow the gap in planning ability between humans and algorithms? In this paper we argue that the answer lies in seeking a formal understanding of how people form cognitive maps to represent predictable structure in the world, and leverage this structure for cost-efficient planning.

\begin{figure}[H]
\begin{center}
\includegraphics[width=9.2cm]{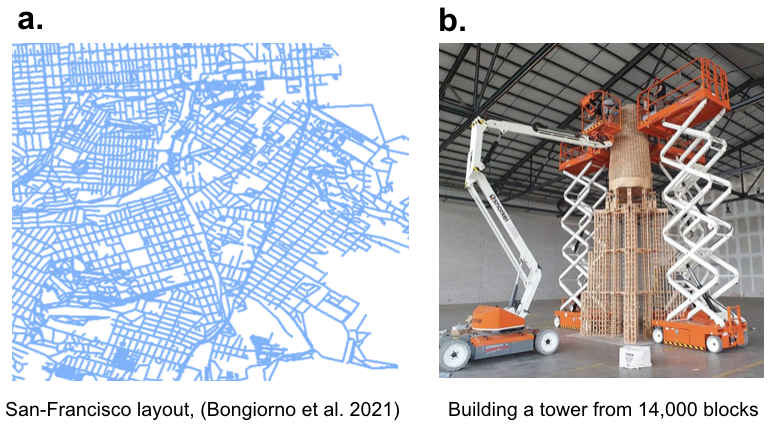}
\includegraphics[width=9.2cm]{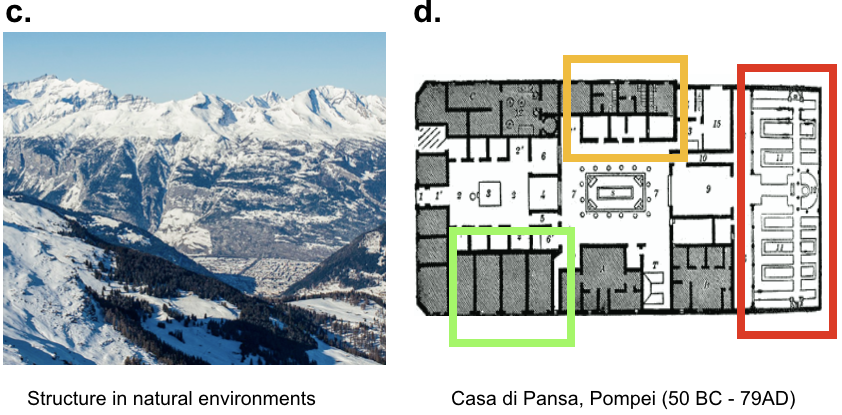}
\end{center}
\caption{Real-world planning tends to occur in predictably structured domains, such as  street networks (a), modular architecture (b,d), and naturally patterned landscapes (c). } 
\label{fig1}
\end{figure}

Current AI formalizes planning as a search within a decision tree of possible actions and outcomes. This tree can be encoded in various ways, such as a learned neural policy~\cite{liu2020experience}, an explicit tree structure~\cite{russell2016artificial}, or a neuro-symbolic model~\cite{tang2024worldcoder}. The size of the underlying decision tree determines the computational cost of the problem, or how difficult it should be. However, the normative difficulty of such models rarely aligns with human experience, as people often solve real-world problems that are theoretically intractable with relative ease. For example, people efficiently navigate cities without knowing exact location of every link in the street network (Fig.~\ref{fig1}a) \cite{bongiorno2021vectorbased}, and plan construction projects with thousands of actions (Fig.~\ref{fig1}b).

This excellent efficiency in real-life planning stands in contrast with often inefficient performance in the laboratory. People deviate from optimal plans in a variety of contexts, including tasks based on multi-arm bandits~\cite{keramati2016adaptive,huys2015interplay}, games, such as chess~\cite{ferreira2013impact}, and many experimental paradigms designed to examine planning behaviors~\cite{unterrainer2004planning, kryven2024approximate,callaway2022rational}. Such deviations from optimality are commonly explained by limiting planning horizon~\cite{ferreira2013impact,kryven2024approximate,van2023expertise} and low-level constraints on perception~\cite{kryven2024approximate}. However, most laboratory-based tasks conspicuously lack the predictable and regular problem structure ubiquitous in real-life, such as hills and valleys in nature (Fig.\ref{fig1}c) and modular built layouts (Fig.\ref{fig1}d). Moreover, classic experimental paradigms often explicitly avoid creating structure in an effort to isolate variables of interest, such as planning depth. 

Here we test the hypothesis that human planning %in natural contexts
leverages prior knowledge about the world, particularly the expectation that the world is highly structured. 
Given this, cognitive maps may be represented as \textit{programs}, % capable of generating potentially infinite maps, 
because programs can capture structures such as symmetries and repeated parts.

Program-structured maps
address a key computational problem:
Planning in partially-observed environments is intractable~\cite{madani2003undecidability}.
Approximations are needed.
Program-structured maps help tackle hard planning problems because %repeated fragments of code allow recycling successful policies across such fragments, essentially decomposing the decision problem.
programs that repeatedly generates identical map fragments allow recycling successful policies, essentially decomposing the decision problem. 
This gives
plans that are locally optimal (within code fragments), but globally suboptimal.
The key idea is to discover fragments of repeated structure in a partially-observed map, and plan policies only once for each repeated fragment. Then, when each fragment is encountered, we avoid costly belief-space planning by reusing previously computed policies.
We probe the extent to which humans similarly recycle successful policies across repeated code fragments, finding that their plans align with this signature of suboptimality.
% We find humans similarly factor their planning efforts and arrive at plans that recycle policies across repeated map structures, giving a fine-grained  suggesting their approximate planning method reflects the same program-like structures our model is attuned for.
%Below, we briefly review relevant work in AI and psychology, followed by a description of our computational methods and experiment design. 
We show that human planning in structured environments is predicted by our generative map model, and can not be explained by alternative models based on unstructured planning with cognitive constrains.

\section{Background}

From young children to hunter-gatherers, people impose structure on the world to solve problems~\cite{pitt2021spatial,lake2020people}. For example, people spontaneously infer correlation between rewards in spatially adjacent locations~\cite{Schulz432534}, and represent spaces hierarchically -- as divided by visible boundaries~\cite{kosslyn1974cognitive}, geography~\cite{stevens1978distortions}, or sub-regions~\cite{hirtle1985evidence}.  
Behavioral evidence for compositional problem representation extends to auditory~\cite{verhoef2014emergence}, visual~\cite{drawgood}, and abstract concept domains~\cite{schulz2017compositional} -- suggesting that the compositional reasoning may be an evolved adaptation to natural structures encountered in daily life~\cite{johnston2022symmetry}.

A recent study found that people form cognitive maps that facilitate planning~\cite{ho2022people}, selectively representing only the parts of the map relevant to goal-directed routes. 
Cognitive maps may also depend on prior expectations about the world, leading people to anticipate regularities in new environments, even when not informed about them in advance \cite{mapinference}. In naturalistic setting adaptive planning draws on complex conceptual prior knowledge of the world, including knowledge of how agents and objects interact, often referred to as \textit{core knowledge} \cite{acquaviva2022communicating,spelke2007coreknowledge,dehaene2006core}.
Learning these natural priors remains an important problem in cognitive AI \cite{kumar2022using, li2024combining, binz2024centaurfoundationmodelhuman}.

Parallel developments in computing have long tackled the goal of building intelligent systems capable of solving general, procedural tasks ~\cite{chollet2025arcprize2024technical,veness2011monte}.
Reinforcement Learning (RL) models abstract problem representations by expressing actions performed together as \textit{options}~\cite{sutton1999between}, learning families of similar Markov Decision Process (MDP) with shared rewards~\cite{wilson2012transfer}, and building efficient state-spaces by recognizing actions that lead to identical observations \cite{singh2012predictive}. Generalized planning frameworks can find algorithm-like policies for solving multiple instances of a task~\cite{curtis2022discovering}, although their ability to handle uncertainty is still limited. A principle of grouping game-board states based on rotation and reflection symmetries was used to optimize representations in the game of Go~\cite{silver2017mastering}, although this work does not consider automatic discovery of structured representations. 

The central challenge to both modeling cognition and engineering general intelligence is therefore learning and leveraging natural priors~\cite{feldman2013tuning}. How can AI learn core knowledge and common sense expectations that make people so efficient in real-life? An emerging line of research leverages LLMs, to make plans in engineering~\cite{tang2024worldcoder} and behavioral domains~\cite{correa2023exploring}.  While the applications of LLMs to directly generating plans are limited, studies have successfully used LLM's program synthesis abilities, to generate programmatic representations of transition function for planning in the OpenAI Gym domain~\cite{tang2024worldcoder,towers2024gymnasium} and to synthesize Planning Domain Definition Language (PDDL) specifications~\cite{xie2023translating}.
Our computational framework uses a similar approach on the assumption that by doing so, we can implicitly access human prior knowledge of the world embedded in language and code used in LLM training.

\begin{figure*}[t]
\begin{center}
\includegraphics[width=\textwidth]{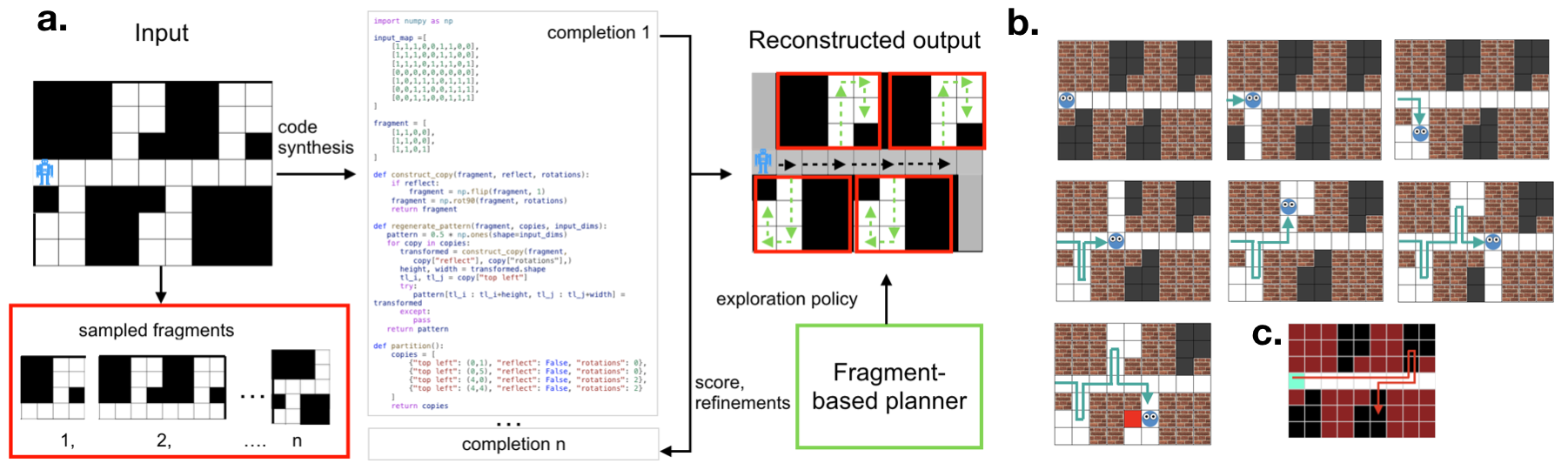}
\end{center}
\caption{ (a) The GMP framework. Generative map representations recovered from the input map constitute a list of fragments, and a program for approximately reconstructing the input from them. The fragments are used by the Planning module to plan a policy for searching each fragment. (b) A series of screens seen by a person in the experiment. Black tiles are initially hidden, and are revealed as they are brought into line of sight. The exit becomes visible in the last screen, shown as a red tile. Arrows indicate the participant's path, searching mazes in a modular way (unit by unit). (c) The optimal path that minimizes the steps needed to reach a randomly placed exit in this maze is non-modular.
} 
\label{fig2}
\end{figure*}

\section{Methods}

\subsection{Behavioral Experiment} 

To test people's natural planning strategies, we adapted a version of Maze Search Task (MST) previously used to evaluate computational models of human planning and plan perception in spatial domains \cite{kryven2024approximate,kryven2021plans}.
The objective of MST is to navigate a series of partially observable, two-dimensional grid-world mazes, finding exits hidden in each maze. As the mazes are partially observable, the exits are initially placed at a random unobserved location within a maze. Fig.\ref{fig2}b shows a sequence of views seen by a participant in MST\footnote{We encourage the reader to try the experiment here: \url{http://18.25.132.241/fragments/int_exp.php}}. People search the mazes by clicking on any unoccupied grid cells adjacent to their character (a round face icon) to move between adjacent cells. The black hidden cells are revealed when they come into the avatar's line of sight.  When revealed, the exit becomes visible as a red tile. As soon as the character moves over the exit, the trial ends.

\subsubsection{Procedure}
Before beginning the experiment people gave informed consent and completed a series of practice mazes, followed by an instruction quiz. Following this, they completed a version of MST with 21 generatively structured mazes, with exit locations randomly chosen at the time of design. In one maze the exit was randomly placed in plain sight, and it was excluded from the analysis. 
After completing MST people completed a Cognitive Reflection Test (CRT) \cite{frederick2005cognitive}, previously shown to correlate with allocating cognitive resources to planning \cite{kryven2021plans}. 
Given that the original CRT was used extensively, our version of CRT provides an analogous set of problems in a novel context~\cite{chandler2014nonnaivete}. Lastly, we administered a post-experiment questionnaire in which people were asked to describe any search strategies they used, and provided demographic information.
As our goal was to observe people's natural planning, we did not offer performance-based incentives. People were informed that the exit could be in any of the hidden tiles, and instructed to find it in each maze.

\subsubsection{Participants} 
We recruited 30 ($13$ female, $17$ male, $M(age)=36,7$, $SD(age)=13.5$) english-speaking participants on Prolific, who were paid 9\pounds per hour. None were excluded. On average the experiment took $10$ minutes to complete. A preliminary pilot experiment revealed a strong effect of modularity, leading us to conclude that a small sample is sufficient to confirm this effect.

\subsection{Computational Model}
\subsubsection{Problem Formulation}

Decision making under partial observability can be modeled by a partially observable Markov decision process (POMDP). Equivalently, it can be viewed as a fully observable search through a space of beliefs, where each belief is a probability distribution over possible states.
Solving POMDPs is notoriously hard~\cite{madani2003undecidability}, hence understanding how people approach these problems holds deep importance for cognitive science and AI.

Formally, a POMDP is a tuple 
\(\langle \Delta(S), A, \tau, r, b_0, \gamma \rangle\),
where \(\Delta(S)\) is the space of probability distributions over a state space \(S\), \(A\) is the set of actions, \(\tau\) is the belief update function, \(r\) is the reward function, \(b_0\) is the initial belief, and \(\gamma\) is the discount factor. The belief state evolves deterministically via \(\tau\), reflecting both the agent's actions and observations.

In this work, each state \(s \in S\) is represented as an \(N \times M\) grid whose cells are labeled \(\{\text{wall}, \text{empty}, \text{exit}, \text{agent}\}\). The overall state space \(S\) consists of all such grids containing exactly one agent and one exit. A belief \(b \in \Delta(S)\) is thus a probability distribution over these grids, encoding the agent’s uncertainty about the true state. Initially, \(b_0\) assumes that the agent and the walls are known, while the exit is uniformly distributed over all valid, unseen cells. The action space \(A\) contains four possible movements (up, down, left, right). Observations \(o \in O\) reveal the visible subset of the grid around the agent, with each visible cell labeled \(\{\text{wall}, \text{empty}, \text{exit}\}\), and any cell outside the agent’s visibility range \(r\) labeled as \emph{unseen}. Observations are consistent with the grid structure of the true state \(s \in S\).

The belief update function \(\tau\) is given by
\[
b'(s') \;\propto\; \; Z(o \mid s') \;\sum_{s \in S} T(s', a, s)\, b(s),
\]
where \(T(s', a, s)\) is the transition function, and \(Z(o \mid s')\) is the observation likelihood. %, and \(\eta\) is a normalizing constant ensuring \(\sum_{s'} b'(s')=1\). 
The transition function \(T(s', a, s)\) specifies the probability of transitioning to state \(s'\) from \(s\) after executing action \(a\). Here, actions that would move the agent into a wall result in the agent remaining in its current position, and transitions to an exit state terminate the process. The observation function \(Z(o \mid s')\) encodes the likelihood of observing \(o\) given \(s'\), where observations reflect the visible subset of the grid within range \(r\) of the agent's position. Visibility is blocked by walls, such that cells beyond a wall are labeled as \emph{unseen}. Finally, the reward function \(r(b, a)\) is the expected reward under the belief \(b\). Since the agent can always see an exit before reaching it, $r(b, a)=1$ if action $a$ leads the agent to a known exit and $0$ otherwise.
\\
\subsubsection{Expected Utility}
The optimal policy for a this POMDP can be found through a belief space tree search~\cite{kaelbling1998planning}. The search is conducted over a tree where each node represents a belief \(b \in \Delta(S)\), and edges correspond to action-observation pairs \((a, o)\). Starting from the root node \(b_0\), the tree expands by simulating actions \(a \in A\) and updating beliefs using the belief update function \(\tau\). For each action \(a\), the agent considers all possible observations \(o \in O\), with the likelihood of each observation determined by the observation function \(Z(o \mid s')\). At each node, the value of a belief is computed recursively using the Bellman equation:
\begin{equation}
V(b) = \max_{a \in A} \left[ r(b, a) + \gamma \sum_{o \in O} P(o \mid b, a) V(\tau(b, a, o)) \right],
\label{eq1}
\end{equation}

where \(P(o \mid b, a)\) is the probability of receiving observation \(o\) after taking action \(a\) under belief \(b\). The optimal policy $\pi^*$ is derived by selecting the action at each belief node that maximizes the expected value. See \cite{kryven2024approximate} for further details on this implementation.

Although this is the optimal strategy, human behavior has previously been shown to diverge at times from its predictions \cite{kryven2024approximate}, where the extent of this divergence varies between individuals in a way that can be explained by the amount of cognitive resources people allocate to planning \cite{kryven2021plans}. Previous work with MST, as well as with related non-spatial planning tasks \cite{huys2015interplay}, has found that people's divergence from the optimal trajectories is most readily explained by a limited planning horizon ( discount factor $\gamma < 1$ in Equation \ref{eq1}).
In the the remainder of this section we describe alternative computational hypotheses for how humans could make decisions in this environment by reasoning about structural patterns.

\begin{algorithm}
\caption{Generative Program from 2D Array}
\label{alg:generative_program}
\begin{algorithmic}[1]
\Require \( I \): Input map, \( t \): Threshold, \( C \): Number of completions
\Ensure \( \lambda \) generative program, \text{fragments}

\State \( S' \gets 0 \)
\State Initialize \( \text{fragments} \gets [] \)
\State \( \lambda \gets "" \)

\While{\( S' < t \)}
    \State Generate a prompt from \( I \) and \( \text{fragments} \)
    \State Send the prompt and receive \( C \) completions
    \State Extract programs \( \{\lambda_1, \lambda_2, \dots, \lambda_C\} \)
    
    \ForAll{\( \lambda_i \in \{\lambda_1, \lambda_2, \dots, \lambda_C\} \)}
        \If{\( \lambda_i \) runs successfully}
            \State \( S_i \gets S(\lambda_i) \)
        \EndIf
    \EndFor
    
    \State \( S', i \gets \max(S_i) \)
    \State \( \lambda \gets \lambda_i \) with the highest score
    \State \( \text{fragments} \gets \text{fragments}_i \) 
\EndWhile

\State \Return \( \lambda \), \text{fragments}
\end{algorithmic}
\end{algorithm}

\subsubsection{Generative Modular Planning (GMP)}
We first describe a model that formalizes planning strategies conditioned on automatically discovered latent structure of the state-space. Our model consists of two modules: a Generative Map Module (GMM) and  Fragment-based Planning (FP) module. See Fig.\ref{fig2}a for a high-level overview of this architecture. The GMM recovers a programmatic representation of the observed state-space, as a composition of fragment units. 
The FP then uses a planner to plan a piece-wise policy once per-fragment, in contrast to a global policy, saving computing costs. 
Importantly, this reconstructed programmatic representation is a cognitively-inspired state-space compression. While such a reconstruction may match the ground-truth planning state-space, it does not need to be exact. In theory, the cognitive principle of combining automatic structure discovery with structure-aware planners can apply to any domain, as a proof of concept here we focus on spatial tasks.

\begin{figure}[t]
\begin{center}
\includegraphics[width=0.9\columnwidth]{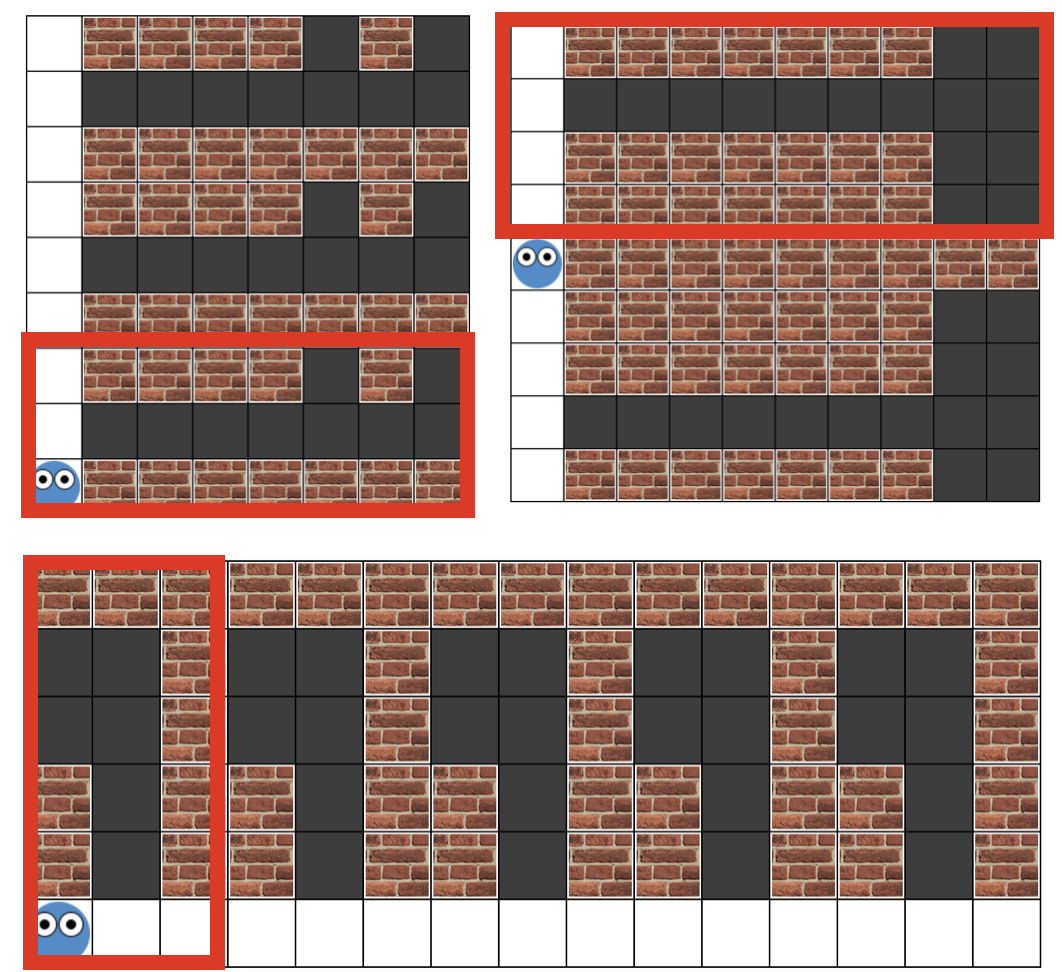}
% panel b and c 
\end{center}
\caption{Examples of maps used in the experiment with 3, 2, and 5 structural fragments highlighted. } 
\label{fig4}
\end{figure}

\noindent
\textbf{Generative Map Module}
We use LLM-based program synthesis with GPT4 to search for programs that can generate cognitive maps based on input (see Algorithm \ref{alg:generative_program}). To do this, we prompt LLM to identify potentially repeating fragments in the input map, and synthesize a Python program that approximately reconstructs the input based on these fragments. The prompt includes Python code with helper functions for admissible fragment transformations, as well as scoring function by which the maps will be evaluated (described below).
In our implementation the input map is a grid-world, specified by an numerical array in Python, where each grid cell is associated with a number (e.g. wall=1, floor=0). The result of this operation is a small number of possible programs that approximately reconstruct the map. This reconstruction allows parts of the output to be left undefined.

We score each candidate map completion by a weighted combination of grid-level similarity and the Minimum Description Length (MDL) principle (\cite{RISSANEN1978465}), to produce a ranking of reconstructed maps. The MDL method penalizes fragments by size as both the shape and each entry must be specified. MDL also penalizes each occurrence used to reconstruct the map by the bits necessary to specify the fragment to map transformation: the translation (as position of top left corner), rotations, and reflection. This means that uninformative fragments (for instance, uniform blocks of cells containing a wall or open space) do not form a parsimonious explanation for the map, even though they of course occur many times. This can be viewed as enforcing a preference for simple generative programs, with the additional structural requirement that \emph{the generative programs use primitives, in this case transformations, encoding symmetries which are relevant to planning.} The highest-ranking map program is then used by the planning module to generate a policy. Formally, the score is
\begin{equation}
    S = \frac{w_1}{M \cdot N} \sum_{x=1}^M \sum_{y=1}^N \left(I(x, y) - O(x, y)\right)^2 - w_2 |\lambda|
    \label{eq2}
\end{equation}

Here $N,M$ are map dimensions, $I$ is the input, $O$ is the output (reconstructed) map, $\lambda$ is the program, and $w_1,w_2$ are weights, free parameters of the model. In the general case, input $I$ and output $O$ are real-valued 2D image arrays. In our current implementation input $I$ takes values $0$ and $1$, and the output $0 \leq O(x,y) \leq 1$. 

Instead of using the raw Python program for $\lambda$ to measure of its complexity, we use a compressed encoding of the fragments, and the transformations used to reconstruct the map. Here compressing LLM-generated fragments and transformations is analogous to refactoring the synthesized programs. As the length of LLM-synthesized code may be noisy, due to injected comments an code redundancies, we refactoring the output obtains a denoised metric of complexity.  %The resulting length is calculated by Algorithm \ref{alg:mdl_score}, where $\overline{n}$ is a self-delimiting (binary) encoding\footnote{The standard binary encoding of $n$ takes $\lceil\lg n\rceil$ bits. To make it self-delimiting we must also encode its length; one way is to pad it with $\lceil\lg n\rceil$ ones terminated by a seperating zero. This approach gives $l(\overline{n}) = 1 + 2 \lceil\lg n\rceil$.} of $n$ and $w,h$ represent width and height respectively. On line \ref{copy_cost} we account for the 
For example, just two bits are necessary to specify a copy's rotation and one bit necessary to specify whether it is reflected. 
%\begin{algorithm}
%%\caption{MDL score}
%\label{alg:mdl_score}
%\begin{algorithmic}[1]
%\Require \( I \): Input map, \( F \): Fragment, \( T \): Transformations
%\Ensure returns \( |\lambda| \)
%\State fragmentCost \(\gets l(\overline{F.w}) + l(\overline{F.h}) + F.w \times F.h\)
%\State costPerCopy \(\gets l(\overline{I.w}) + l(\overline{I.h}) + 2 + 1\)
%\State copyCost \(\gets l(\overline{\# T}) + \# T \times \) costPerCopy \label{copy_cost}
%\State omissions \(\gets \# \text{cells in I not covered by T} \)
%\State \Return fragmentCost + copyCost + omissions
%\end{algorithmic}
%\end{algorithm}

%The full prompt text will be available in SI with the camera-ready version of this paper. % TBD COME BACK TO THIS IF HAVE TIME
Fig.\ref{fig4} shows examples of three maps used in the experiment with 4, 2 and 5 structural fragments, shown highlighted in red.

\subsubsection{Fragment-based Planning Module} 
In the current implementation we adapt the optimal Expected Utility model (an implementation of $\pi^*$) from \cite{kryven2024approximate} to plan within fragments. %Given a relatively simple nature of the fragments used in our current experiment, people  search the fragments optimally in alignment with this model. 
The implication of this modeling choice is to assume that people are locally optimal but globally suboptimal, as is a natural consequence of a model that can exactly solve small problems and then reuse them.
To plan between fragments (in parts of the output left undefined) FP moves toward the closest fragment by locally solving a Markov Decision Process (using value iteration), on the assumption that the prior probability of finding the exit is uniform across fragments.
When such priors are non-uniform, the problem of planning between fragments can be solved by any planning algorithm.

This approach is resource-efficient, as our planner computes a decision (sub)tree only once per fragment, and subsequently reuses it every time the fragment is encountered. Instead of computing a decision-tree for the entire map, our model maintains several smaller decision-trees, %one for each way of entering each type of fragment within a map, 
along with a generative program that describes how the global map can be reconstructed from them. 
In principle, as the map reconstruction is approximate the planner may encounter observations inconsistent with the reconstructed map. When this happens, revert to non-modular planning (e.g. using the Expected Utility planning~\cite{kryven2024approximate}) to ensure robustness of the algorithm.
While our choice of planner was motivated by comparing results to prior work, our framework does not critically depend on the choice of the internal non-modular planner, and can integrate with other implementations.

\subsubsection{Alternative models} We compare GMP to an existing set of models previously used to explain human behavior in MST~\cite{kryven2024approximate}. The models treat maze search as a path minimization problem - an implicit goal spontaneously reported by participants in the post-experiment questionnaire. Here we compared people to a subset of models from  pervious work -- the Expected Utility (optimal planner) and the Discounted Utility (an overall best-fitting planner, assuming $\gamma = 0.7$ which plans with a limited horizon)\cite{kryven2024approximate}. To facilitate this comparison, we designed the environments so that these existing models either predict non-modular policies, or are indifferent between modular and non-modular search. Whenever an existing model is indifferent between modular and non-modular paths in an environment, the probability of such a model taking a modular path is at most 0.25.

\section{Results}
\begin{figure}[t]
\begin{center}
\includegraphics[width=\columnwidth]{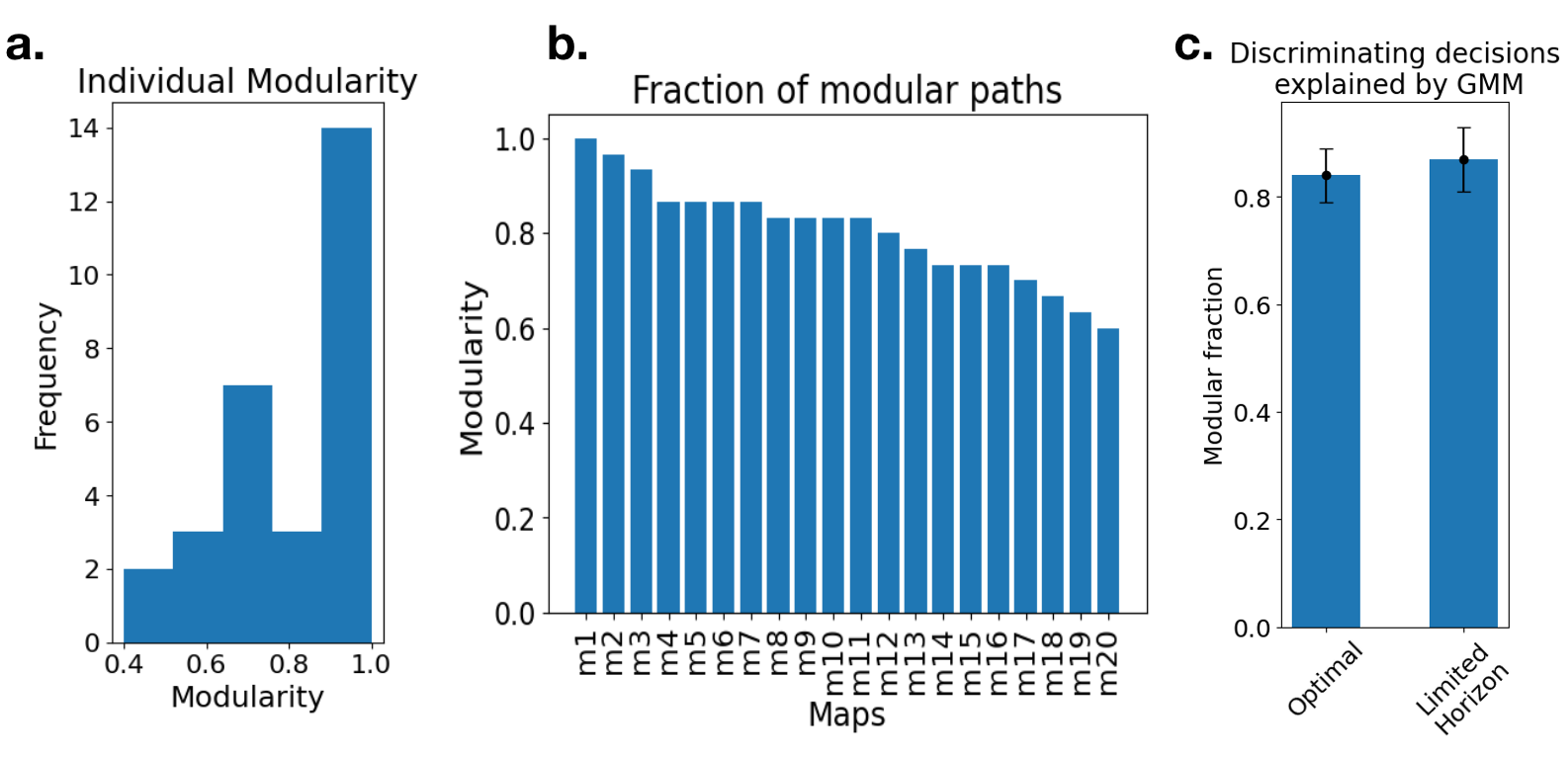}
% panel b and c 
\end{center}
\caption{ People are highly modular, according to the conservative definition. (a) Histogram of modularity across participants.  (b) %The relationship between people's modularity and their CRT score shows that people who allocate more cognitive resources make less modular plans.
Fraction of people who searched using modular trajectories, for each of the 20 maps. (c) Aggregating across individual decisions and people. Each bar shows the fraction of discriminating decisions per participant where a participant's choice was predicted by our model, but and not by the alternative model. Error bars are 95\% CI over people.
} 
\label{fig3}
\end{figure}

We introduce the following metrics for comparing behavior to our models.
First, we define a conservative definition of a \textbf{modular path} as a path that visits all fragments in order, always moving to the closest subsequent fragment. This definition of modularity is consistent with our implementation of GMP, however, it underestimates the true rates of modularity since people who represent a map as modular may have non-uniform prior beliefs about which fragments are 'rewarding'. For example, people could assume that the exit tends to be in the second or third fragment, skipping some parts of the map. 
We use this definition to compute  \textbf{modularity} -- that is, the fraction modular paths for each individual (Fig.\ref{fig3}a) and for each environment (Fig.\ref{fig3}b). 
Second, we define as a \textbf{discriminating decision} as any state of the environment in which GMM and an alternative model predict a different most likely path. As different models can predict different paths, the set of discriminating decisions will different for any given pair of models, given the same set of environments. By measuring how well the alternative models predict people in discriminating decisions, we account for the possibility that people could be using several planning strategies within one environment, including switching between Modular and non-modular strategies. 

Examination of individual and map-specific modularities in Figures \ref{fig3}a and \ref{fig3}b shows that people are highly consistent with our model, demonstrating signatures of modular planning across all environments and participants.
In contrast, by our experiment design the alternative models predict that the mazes should be searched in a non-modular way.
Further, GMM predicts people significantly better than the optimal planner, or planning with limited horizon perviously shown explain human behavior in MST (Fig.\ref{fig3}). 

% TBD TEST IF THIS HOLDS?
To test the hypothesis that modular planning is a resource-efficient adaptation for reducing cognitive costs, we compute a linear regression of people's CRT score against their modularity rate. %CRT scores were previously shown to correlate with the allocation of cognitive resources to planning \cite{kryven2021plans}, a negative correlation between these scores would support our hypothesis. 
While we found a negative correlation between these scores during pilots, it was not significant in the current results.% ($p = 0.4$). 
This lack of replication is likely due to our conservative definition of modularity underestimating the true rate whenever people searching fragments in a different order. 

%\section{Acknowledgments}

%funding information, anyone contributing to the discussion.

\section{Discussion}

% summary of what we did
We test the hypothesis that people may be representing cognitive maps by generative programs, conditioned on conceptual prior knowledge of the world. We leverage an experimental task previously used to study human planning, with partially observable environment layouts designed to differentiate between Modular planning, planning that follows optimally computed Expected Utility, and the model previously shown to describe planning by a limited horizon. 
We found that people follow highly modular strategies, that generalize across environments and individuals in support of our hypothesis.
We interpret our results by a computational model that combines LLM program synthesis for map generation with modular planning, showing that this model is highly accurate at predicting behavior. This result suggests  that people deviate from the optimal plans in structured environments at least in part due to reasoning about the environmental structure.

% contributions
We makes three contributions in the scientific and engineering domains. First, we make a scientific contribution, showing that people make adaptive plans consistent with program-like representations of cognitive maps.
Second, we advance the understanding of latent cognitive abilities of LLMs by showing how to elicit human inductive biases about spatial decision-making.
Third, we contribute an implementation showing how to actually build these principles into a working system.
Our GMP model differs from existing hierarchical models (which divide a complex problem into manageable chunks), as it can additionally reuse computations across repeating fragments. It also differs from existing generalizable task-and-motion models specified in PDDL, as it tolerates uncertainty.
By using LLMs for program-synthesis, our work is more practical and scalable than the traditional enumerative search (brute-force) methods 
~\cite{mapinference,veness2011monte} for discovery of structural form.

Notably, we find variability in behavior between individuals and between maps that is not accounted for by our conservative definition of Modular planning. Much of this variability arises from people skipping fragments, suggesting that future versions of the model should include non-uniform priors over which fragments people believe to be rewarding. Future studied should also consider how latent cognitive factors, such as learning, attention, and available cognitive resources, may affect how people modularize and represent maps. 

% everything we did not do
As map reconstructions are approximate, the planner may encounter observations inconsistent with the reconstructed map. We currently handle such scenarios by  reverting to a non-modular planner, however the alignment of this solution with people remains to be evaluated experimentally. More studies are needed to understand how people process discrepancies between their cognitive map and reality, when their expectations fail.
Future studies should also examine the extent of variation between fragments perceived as instances of the same class in the light of possible goals, as cognitive maps are likely to be goal-dependent~\cite{ho2022people}.

While planning cognition has been studied extensively, the domain of real-world planning remains underexplored. Our work contributes a formal computational insight into how human prior knowledge may guide resource-efficient plans, as well as contributes to the emerging investigation of whether LLMs can capture naturalistic inductive biases (e.g., \cite{tang2024worldcoder}), offering insights into this question in the spatial domain.
By expressing our findings in computational terms, our GMP model moves toward translating cognitive mechanisms behind human planning into AI applications.

%\bibliography{CogSci_Template}
%\printbibliography

%\bibliography{CogSci_Template}

\begin{thebibliography}{}

\bibitem[Acquaviva et~al., 2022]{acquaviva2022communicating}
Acquaviva, S., Pu, Y., Kryven, M., Sechopoulos, T., Wong, C., Ecanow, G., Nye, M., Tessler, M., and Tenenbaum, J. (2022).
\newblock Communicating natural programs to humans and machines.
\newblock {\em Advances in Neural Information Processing Systems}, 35:3731--3743.

\bibitem[Binz et~al., 2024]{binz2024centaurfoundationmodelhuman}
Binz, M., Akata, E., Bethge, M., Brändle, F., Callaway, F., Coda-Forno, J., Dayan, P., Demircan, C., Eckstein, M.~K., Éltető, N., Griffiths, T.~L., Haridi, S., Jagadish, A.~K., Ji-An, L., Kipnis, A., Kumar, S., Ludwig, T., Mathony, M., Mattar, M., Modirshanechi, A., Nath, S.~S., Peterson, J.~C., Rmus, M., Russek, E.~M., Saanum, T., Scharfenberg, N., Schubert, J.~A., Buschoff, L. M.~S., Singhi, N., Sui, X., Thalmann, M., Theis, F., Truong, V., Udandarao, V., Voudouris, K., Wilson, R., Witte, K., Wu, S., Wulff, D., Xiong, H., and Schulz, E. (2024).
\newblock Centaur: a foundation model of human cognition.

\bibitem[Bongiorno et~al., 2021]{bongiorno2021vectorbased}
Bongiorno, C., Zhou, Y., Kryven, M., Theurel, D., Rizzo, A., Santi, P., Tenenbaum, J., and Ratti, C. (2021).
\newblock Vector-based pedestrian navigation in cities.

\bibitem[Callaway et~al., 2022]{callaway2022rational}
Callaway, F., van Opheusden, B., Gul, S., Das, P., Krueger, P.~M., Griffiths, T.~L., and Lieder, F. (2022).
\newblock Rational use of cognitive resources in human planning.
\newblock {\em Nature Human Behaviour}, 6(8):1112--1125.

\bibitem[Chandler et~al., 2014]{chandler2014nonnaivete}
Chandler, J., Mueller, P., and Paolacci, G. (2014).
\newblock Nonnaivete among amazon mechanical turk workers: Consequences and solutions for behavioral researchers.
\newblock {\em Behavior research methods}, 46(1):112--130.

\bibitem[Chollet et~al., 2025]{chollet2025arcprize2024technical}
Chollet, F., Knoop, M., Kamradt, G., and Landers, B. (2025).
\newblock Arc prize 2024: Technical report.

\bibitem[Correa et~al., 2023]{correa2023exploring}
Correa, C.~G., Sanborn, S., Ho, M.~K., Callaway, F., Daw, N.~D., and Griffiths, T.~L. (2023).
\newblock Exploring the hierarchical structure of human plans via program generation.
\newblock {\em arXiv preprint arXiv:2311.18644}.

\bibitem[Curtis et~al., 2022]{curtis2022discovering}
Curtis, A., Silver, T., Tenenbaum, J.~B., Lozano-P{\'e}rez, T., and Kaelbling, L. (2022).
\newblock Discovering state and action abstractions for generalized task and motion planning.
\newblock In {\em Proceedings of the AAAI conference on artificial intelligence}, volume~36, pages 5377--5384.

\bibitem[Dehaene et~al., 2006]{dehaene2006core}
Dehaene, S., Izard, V., Pica, P., and Spelke, E. (2006).
\newblock Core knowledge of geometry in an amazonian indigene group.
\newblock {\em Science}, 311(5759):381--384.

\bibitem[Feldman, 2013]{feldman2013tuning}
Feldman, J. (2013).
\newblock Tuning your priors to the world.
\newblock {\em Topics in cognitive science}, 5(1):13--34.

\bibitem[Ferreira, 2013]{ferreira2013impact}
Ferreira, D.~R. (2013).
\newblock The impact of the search depth on chess playing strength.
\newblock {\em ICGA journal}, 36(2):67--80.

\bibitem[Frederick, 2005]{frederick2005cognitive}
Frederick, S. (2005).
\newblock Cognitive reflection and decision making.
\newblock {\em The Journal of Economic Perspectives}, 19(4):25--42.

\bibitem[Hirtle and Jonides, 1985]{hirtle1985evidence}
Hirtle, S.~C. and Jonides, J. (1985).
\newblock Evidence of hierarchies in cognitive maps.
\newblock {\em Memory \& cognition}, 13(3):208--217.

\bibitem[Ho et~al., 2022]{ho2022people}
Ho, M.~K., Abel, D., Correa, C.~G., Littman, M.~L., Cohen, J.~D., and Griffiths, T.~L. (2022).
\newblock People construct simplified mental representations to plan.
\newblock {\em Nature}, 606(7912):129--136.

\bibitem[Huys et~al., 2015]{huys2015interplay}
Huys, Q.~J., Lally, N., Faulkner, P., Eshel, N., Seifritz, E., Gershman, S.~J., Dayan, P., and Roiser, J.~P. (2015).
\newblock Interplay of approximate planning strategies.
\newblock {\em Proceedings of the National Academy of Sciences}, 112(10):3098--3103.

\bibitem[Johnston et~al., 2022]{johnston2022symmetry}
Johnston, I.~G., Dingle, K., Greenbury, S.~F., Camargo, C.~Q., Doye, J.~P., Ahnert, S.~E., and Louis, A.~A. (2022).
\newblock Symmetry and simplicity spontaneously emerge from the algorithmic nature of evolution.
\newblock {\em Proceedings of the National Academy of Sciences}, 119(11):e2113883119.

\bibitem[Kaelbling et~al., 1998]{kaelbling1998planning}
Kaelbling, L.~P., Littman, M.~L., and Cassandra, A.~R. (1998).
\newblock Planning and acting in partially observable stochastic domains.
\newblock {\em Artificial intelligence}, 101(1-2):99--134.

\bibitem[Keramati et~al., 2016]{keramati2016adaptive}
Keramati, M., Smittenaar, P., Dolan, R.~J., and Dayan, P. (2016).
\newblock Adaptive integration of habits into depth-limited planning defines a habitual-goal--directed spectrum.
\newblock {\em Proceedings of the National Academy of Sciences}, 113(45):12868--12873.

\bibitem[Kosslyn et~al., 1974]{kosslyn1974cognitive}
Kosslyn, S.~M., Pick~Jr, H.~L., and Fariello, G.~R. (1974).
\newblock Cognitive maps in children and men.
\newblock {\em Child development}, pages 707--716.

\bibitem[Kryven et~al., 2021]{kryven2021plans}
Kryven, M., Ullman, T.~D., Cowan, W., and Tenenbaum, J.~B. (2021).
\newblock Plans or outcomes: How do we attribute intelligence to others?
\newblock {\em Cognitive Science}, 45(9):13--41.

\bibitem[Kryven et~al., 2024]{kryven2024approximate}
Kryven, M., Yu, S., Kleiman-Weiner, M., Ullman, T., and Tenenbaum, J. (2024).
\newblock Approximate planning in spatial search.
\newblock {\em PLOS Computational Biology}, 20(11):e1012582.

\bibitem[Kumar et~al., 2022]{kumar2022using}
Kumar, S., Correa, C.~G., Dasgupta, I., Marjieh, R., Hu, M., Hawkins, R.~D., Cohen, J., Daw, N., Narasimhan, K.~R., and Griffiths, T.~L. (2022).
\newblock Using natural language and program abstractions to instill human inductive biases in machines.
\newblock In Oh, A.~H., Agarwal, A., Belgrave, D., and Cho, K., editors, {\em Advances in Neural Information Processing Systems}.

\bibitem[Lake and Piantadosi, 2020]{lake2020people}
Lake, B.~M. and Piantadosi, S.~T. (2020).
\newblock People infer recursive visual concepts from just a few examples.
\newblock {\em Computational Brain \& Behavior}, 3(1):54--65.

\bibitem[Li et~al., 2024]{li2024combining}
Li, W.-D., Hu, K., Larsen, C., Wu, Y., Alford, S., Woo, C., Dunn, S.~M., Tang, H., Naim, M., Nguyen, D., et~al. (2024).
\newblock Combining induction and transduction for abstract reasoning.
\newblock {\em arXiv preprint arXiv:2411.02272}.

\bibitem[Liu et~al., 2020]{liu2020experience}
Liu, Y., Mattar, M.~G., Behrens, T.~E., Daw, N.~D., and Dolan, R.~J. (2020).
\newblock Experience replay supports non-local learning.
\newblock {\em BioRxiv}, pages 2020--10.

\bibitem[Madani et~al., 2003]{madani2003undecidability}
Madani, O., Hanks, S., and Condon, A. (2003).
\newblock On the undecidability of probabilistic planning and related stochastic optimization problems.
\newblock {\em Artificial Intelligence}, 147(1-2):5--34.

\bibitem[Pitt et~al., 2021]{pitt2021spatial}
Pitt, B., Ferrigno, S., Cantlon, J.~F., Casasanto, D., Gibson, E., and Piantadosi, S.~T. (2021).
\newblock Spatial concepts of number, size, and time in an indigenous culture.
\newblock {\em Science Advances}, 7(33):eabg4141.

\bibitem[Rissanen, 1978]{RISSANEN1978465}
Rissanen, J. (1978).
\newblock Modeling by shortest data description.
\newblock {\em Automatica}, 14(5):465--471.

\bibitem[Russell and Norvig, 2016]{russell2016artificial}
Russell, S.~J. and Norvig, P. (2016).
\newblock {\em Artificial intelligence: a modern approach}.
\newblock Pearson.

\bibitem[Schulz et~al., 2018]{Schulz432534}
Schulz, E., Franklin, N.~T., and Gershman, S.~J. (2018).
\newblock Finding structure in multi-armed bandits.
\newblock {\em bioRxiv}.

\bibitem[Schulz et~al., 2017]{schulz2017compositional}
Schulz, E., Tenenbaum, J.~B., Duvenaud, D., Speekenbrink, M., and Gershman, S.~J. (2017).
\newblock Compositional inductive biases in function learning.
\newblock {\em Cognitive psychology}, 99:44--79.

\bibitem[Sharma et~al., 2022]{mapinference}
Sharma, S., Curtis, A., Kryven, M., Tenenbaum, J., and Fiete, I. (2022).
\newblock Map induction: Compositional spatial submap learning for efficient exploration in novel environments.
\newblock {\em International Conference of Learning Representations}.

\bibitem[Silver et~al., 2017]{silver2017mastering}
Silver, D., Schrittwieser, J., Simonyan, K., Antonoglou, I., Huang, A., Guez, A., Hubert, T., Baker, L., Lai, M., Bolton, A., et~al. (2017).
\newblock Mastering the game of go without human knowledge.
\newblock {\em nature}, 550(7676):354--359.

\bibitem[Singh et~al., 2012]{singh2012predictive}
Singh, S., James, M., and Rudary, M. (2012).
\newblock Predictive state representations: A new theory for modeling dynamical systems.
\newblock {\em arXiv preprint arXiv:1207.4167}.

\bibitem[Spelke and Kinzler, 2007]{spelke2007coreknowledge}
Spelke, E.~S. and Kinzler, K.~D. (2007).
\newblock Core knowledge.
\newblock {\em Developmental Science}, 10(1):89--96.

\bibitem[Stevens and Coupe, 1978]{stevens1978distortions}
Stevens, A. and Coupe, P. (1978).
\newblock Distortions in judged spatial relations.
\newblock {\em Cognitive psychology}, 10(4):422--437.

\bibitem[Sutton et~al., 1999]{sutton1999between}
Sutton, R.~S., Precup, D., and Singh, S. (1999).
\newblock Between mdps and semi-mdps: A framework for temporal abstraction in reinforcement learning.
\newblock {\em Artificial intelligence}, 112(1-2):181--211.

\bibitem[Tang et~al., 2024]{tang2024worldcoder}
Tang, H., Key, D., and Ellis, K. (2024).
\newblock Worldcoder, a model-based llm agent: Building world models by writing code and interacting with the environment.
\newblock {\em arXiv preprint arXiv:2402.12275}.

\bibitem[Tian et~al., 2020]{drawgood}
Tian, L., Ellis, K., Kryven, M., and Tenenbaum, J. (2020).
\newblock Learning abstract structure for drawing by efficient motor program induction.
\newblock {\em Advances in Neural Information Processing Systems}, 33:2686--2697.

\bibitem[Towers et~al., 2024]{towers2024gymnasium}
Towers, M., Kwiatkowski, A., Terry, J., Balis, J.~U., De~Cola, G., Deleu, T., Goul{\~a}o, M., Kallinteris, A., Krimmel, M., KG, A., et~al. (2024).
\newblock Gymnasium: A standard interface for reinforcement learning environments.
\newblock {\em arXiv preprint arXiv:2407.17032}.

\bibitem[Unterrainer et~al., 2004]{unterrainer2004planning}
Unterrainer, J.~M., Rahm, B., Kaller, C.~P., Leonhart, R., Quiske, K., Hoppe-Seyler, K., Meier, C., M{\"u}ller, C., and Halsband, U. (2004).
\newblock Planning abilities and the tower of london: is this task measuring a discrete cognitive function?
\newblock {\em Journal of clinical and experimental neuropsychology}, 26(6):846--856.

\bibitem[van Opheusden et~al., 2023]{van2023expertise}
van Opheusden, B., Kuperwajs, I., Galbiati, G., Bnaya, Z., Li, Y., and Ma, W.~J. (2023).
\newblock Expertise increases planning depth in human gameplay.
\newblock {\em Nature}, pages 1000--1005.

\bibitem[Veness et~al., 2011]{veness2011monte}
Veness, J., Ng, K.~S., Hutter, M., Uther, W., and Silver, D. (2011).
\newblock A monte-carlo aixi approximation.
\newblock {\em Journal of Artificial Intelligence Research}, 40:95--142.

\bibitem[Verhoef et~al., 2014]{verhoef2014emergence}
Verhoef, T., Kirby, S., and De~Boer, B. (2014).
\newblock Emergence of combinatorial structure and economy through iterated learning with continuous acoustic signals.
\newblock {\em Journal of Phonetics}, 43:57--68.

\bibitem[Wilson et~al., 2012]{wilson2012transfer}
Wilson, A., Fern, A., and Tadepalli, P. (2012).
\newblock Transfer learning in sequential decision problems: A hierarchical bayesian approach.
\newblock In {\em Proceedings of ICML Workshop on Unsupervised and Transfer Learning}, pages 217--227. JMLR Workshop and Conference Proceedings.

\bibitem[Xie et~al., 2023]{xie2023translating}
Xie, Y., Yu, C., Zhu, T., Bai, J., Gong, Z., and Soh, H. (2023).
\newblock Translating natural language to planning goals with large-language models.
\newblock {\em arXiv preprint arXiv:2302.05128}.

\end{thebibliography}
\end{document}